\documentclass[conference]{IEEEtran}
\IEEEoverridecommandlockouts
\usepackage{cite}
\usepackage{amsmath,amssymb,amsfonts}
\usepackage{algorithmic}
\usepackage{graphicx}
\usepackage{booktabs,tabularx}
\usepackage{textcomp}
 \usepackage{booktabs}
\usepackage{xcolor}
\def\BibTeX{{\rm B\kern-.05em{\sc i\kern-.025em b}\kern-.08em
    T\kern-.1667em\lower.7ex\hbox{E}\kern-.125emX}}
\begin{document}

\renewcommand{\thefootnote}{\arabic{footnote}}

\newcommand{\equalcontrib}{\textsuperscript{$^{*}$}}
\newcommand{\corrauthor}{\textsuperscript{\ddag}}

\title{Silhouette-to-Contour Registration: Aligning Intraoral Scan Models with Cephalometric Radiographs
\thanks{\noindent\parbox{\linewidth}{%
\equalcontrib~Co-first authors.\\
\corrauthor~Corresponding authors.}}}

\author{%
  \IEEEauthorblockN{%
    Yiyi Miao\textsuperscript{1,2}\equalcontrib,
    Taoyu Wu\textsuperscript{3,4}\equalcontrib,
    Ji Jiang\textsuperscript{5},
    Tong Chen\textsuperscript{1,2},
    Zhe Tang\textsuperscript{6}, \\
    Zhengyong Jiang\textsuperscript{1},
    Angelos Stefanidis\textsuperscript{1},
    Limin Yu\textsuperscript{3}\corrauthor,
    Jionglong Su\textsuperscript{1}\corrauthor}
  \IEEEauthorblockA{%
    \textsuperscript{1}School of AI and Advanced Computing, Xi'an Jiaotong-Liverpool University, China\\
    \textsuperscript{2}School of Electrical Engineering, Electronics and Computer Science, University of Liverpool, United Kingdom\\
    \textsuperscript{3}School of Advanced Technology, Xi'an Jiaotong-Liverpool University, China\\
    \textsuperscript{4}School of Physical Sciences, University of Liverpool, Liverpool, United Kingdom\\
    \textsuperscript{5}School of Mathematics and Physics, Xi'an Jiaotong-Liverpool University, China\\
    \textsuperscript{6}Institute of Artificial Intelligence Innovation, Zhejiang University of Technology, China}
}

\maketitle

\begin{abstract}
Reliable 3D--2D alignment between intraoral scan (IOS) models and lateral cephalometric radiographs is critical for orthodontic diagnosis, yet conventional intensity-driven registration methods struggle under real clinical conditions, where cephalograms exhibit projective magnification, geometric distortion, low-contrast dental crowns, and acquisition-dependent variation. These factors hinder the stability of appearance-based similarity metrics and often lead to convergence failures or anatomically implausible alignments. To address these limitations, we propose \textbf{DentalSCR}, a pose-stable, contour-guided framework for accurate and interpretable silhouette-to-contour registration. Our method first constructs a U-Midline Dental Axis (UMDA) to establish a unified cross-arch anatomical coordinate system, thereby stabilizing initialization and standardizing projection geometry across cases. Using this reference frame, we generate radiograph-like projections via a surface-based DRR formulation with coronal-axis perspective and Gaussian splatting, which preserves clinical source--object--detector magnification and emphasizes external silhouettes. Registration is then formulated as a 2D similarity transform optimized with a symmetric bidirectional Chamfer distance under a hierarchical coarse-to-fine schedule, enabling both large capture range and subpixel-level contour agreement. We evaluate DentalSCR on 34 expert-annotated clinical cases. Experimental results demonstrate substantial reductions in landmark error—particularly at posterior teeth—tighter dispersion on the lower jaw, and low Chamfer and controlled Hausdorff distances at the curve level. These findings indicate that DentalSCR robustly handles real-world cephalograms and delivers high-fidelity, clinically inspectable 3D--2D alignment, outperforming conventional baselines.
\end{abstract}

\begin{IEEEkeywords}
Intraoral scan, Cephalometric Radiograph, 3D–2D registration
\end{IEEEkeywords}

\section{Introduction}

The lateral cephalometric radiograph (CR) is pivotal in orthodontic diagnosis and treatment planning. The integration of 3D dental models into the cephalometric coordinate system is increasingly required for reproducible measurements and consistent 3D–2D visualization~\cite{maintz1998survey}~\cite{zitova2003image}. However, clinical X-rays exhibit projective magnification and geometric distortion from the point source, making direct orthogonal projection inaccurate. Furthermore, dental crown contours in lateral projections are thin and low-contrast, susceptible to artifacts and occlusion, which compromises registration robustness. Conventional intensity-driven 2D–3D registration methods depend on accurate forward imaging models and similarity metrics like mutual information, but they typically have limited convergence range under exposure variations and scatter~\cite{wells1996multi}. This creates a clear clinical need for a geometry-based approach that is interpretable, robust to initialization, and suitable for diagnostic use~\cite{penney1998comparison}.

Existing methodologies comprise three main categories. Intensity-driven methods synthesize DRR and maximize similarity with acquired X-rays~\cite{maes2002multimodality}, yet they remain sensitive to real-world acquisition variations with limited capture range~\cite{viola1997alignment}. Geometry-driven methods utilize contours or edges as anchors via distance fields or point set registration~\cite{jiang1992image}, but face instability in thin, low-contrast structures. Hybrid and differentiable rendering techniques unify appearance and geometry~\cite{swinehart1962beer}, albeit with increased complexity. While X-ray imaging physics provides reliable geometric constraints~\cite{rino2013evaluation}, a practical solution incorporating structural priors with a stable, interpretable objective function remains lacking~\cite{levine2020drrgenerator}.

To address this, we propose \emph{DentalSCR}, a 3D–2D dental model registration framework grounded in an "analysis-by-synthesis" paradigm with a contour-distance field at its core. The method begins by constructing a U-midline Dental Axis (UMDA) to establish a shared and reproducible anatomical coordinate system between the upper and lower jaws, standardizing pose initialization and midline consistency while mitigating the impact of inter-individual variation on optimization. It subsequently employs perspective projection and cephalometry-like Gaussian splatting rendering to explicitly preserve the magnification and distortion inherent to the source-object-detector geometry, thereby ensuring geometric alignment between synthesized projections and clinical cephalograms; the geometric correctness can be cross-validated using open-source Digitally Reconstructed Radiography (DRR) tools when necessary~\cite{unberath2018deepdrr}. The registration process itself is formulated as the geometric matching of projected STL contours to expert-annotated cephalogram contours, minimizing a symmetric Chamfer distance field loss under 2D similarity transformation through a three-stage coarse-to-fine schedule designed to expand the capture range and achieve sub-pixel refinement within the convergence neighborhood. This is complemented by Hausdorff distance metrics to diagnose residual outliers and local worst-case misalignments~\cite{huttenlocher2002comparing}. This design maintains geometric interpretability while circumventing any reliance on complex modeling of illumination and contrast, and remains compatible with—and allows for the substitution of—common point-set optimization tools such as Iterative Closest Point (ICP)~\cite{besl1992method} and Coherent Point Drift (CPD)~\cite{myronenko2010point}.

Our contributions are threefold. First, we introduce the \emph{U-Midline Dental Axis} (UMDA), a cross-arch reference frame that standardizes initialization and evaluation coordinates, thereby stabilizing geometric optimization. Second, we model clinical acquisition with a coronal-axis perspective and a surface-based DRR with radiograph-like splatting, narrowing the geometric gap between synthesized and acquired images within an analysis-by-synthesis framework. \textit{Third}, we cast silhouette-to-contour registration as a symmetric distance-field objective based on bidirectional Chamfer distance, optimized with a hierarchical coarse-to-fine schedule that balances capture range and subpixel accuracy. We validate DentalSCR on 34 expert-annotated clinical cases: upper-jaw landmark mean error and RMSE decrease substantially, dispersion on the lower jaw tightens, and the largest gains appear in posterior teeth. Curve-level metrics further show low bidirectional Chamfer distances and controlled Hausdorff tails, with residuals concentrated in thin, high-curvature regions. Consistent with these findings, recent systematic reviews and multi-center evidence on automated cephalometric landmarking support the clinical plausibility and translational potential of our approach (e.g., Serafin et al.~\cite{serafin2023accuracy} and Hendrickx et al.~\cite{hendrickx2024can}), while studies on arch shape and anatomical priors provide additional methodological context~\cite{lee2024comparative}.

\section{Related Work}
\subsection{Paradigms of 2D–3D Registration in Medical Imaging}

In medical imaging, 2D–3D registration is broadly categorized into three paradigms: intensity-based, feature/geometry-based, and hybrid methods. Intensity-based approaches optimize a similarity metric objective function between the acquired X-ray image and a Digitally Reconstructed Radiograph (DRR) synthesized from a 3D volume or surface model~\cite{maintz1998survey}. Metrics such as Mutual Information and Normalized Cross-Correlation are most prevalent, valued for their robustness in cross-modal scenarios or under varying contrast conditions~\cite{zitova2003image}. This paradigm is contingent upon a physically plausible forward imaging model and computationally efficient DRR generation strategies, and is typically coupled with multi-resolution or coarse-to-fine optimization schemes to enlarge the capture range~\cite{penney1998comparison}. Review articles have systematically consolidated this trajectory and its subsequent variants.

Feature or geometry-driven methods use inter-modally stable structures as anchors (landmarks, curves, contours, edges, or silhouettes) and achieve registration through point-to-set or set-to-set geometric distances. Recent advances have progressed in three key aspects: First, for curve or boundary alignment, Chamfer distance transform metrics~\cite{sun2024medical} continue to be adopted, while enhancements in distance field construction and mask design improve numerical stability and sub-pixel resolution. Second, for explicit correspondence-based rigid/near-rigid alignment, various robust or adaptively weighted ICP (Iterative Closest Point)~\cite{yang2024accurate} variants have been proposed to handle outliers, low overlap, and noise. Third, in learned registration and neural field frameworks, differentiable geometric losses are introduced as supervision to enhance cross-modal generalization and optimization differentiability. These advancements make geometry-based methods more robust under varying imaging devices, artifacts, and exposure changes, particularly suitable for contour/silhouette alignment in X-ray scenarios.

Hybrid methods integrate appearance evidence with geometric constraints within a unified objective, or alternate between contour-driven pose refinement and intensity-driven fine-tuning. Recent trends involve leveraging differentiable X-ray rendering/physical imaging models to incorporate both DRR intensity similarity and geometric terms (e.g., silhouettes, boundaries, keypoints, or regularizations) into end-to-end or two-stage optimization frameworks~\cite{gopalakrishnan2024intraoperative}. Simultaneously, techniques such as coarse-to-fine parameter scheduling, robust penalty functions, and normalized initialization are employed to enlarge the convergence domain. Emerging research from computer vision and medical imaging demonstrates that such approaches significantly enhance the stability and accuracy of 2D-3D registration under multi-view, cross-patient generalization, and weakly-supervised conditions~\cite{leskovar2025comparison}.

\subsection{Structural Priors for 3D Dental Geometry}
3D dental models come with strong, actionable priors that can be encoded for registration and reconstruction~\cite{miao2025dentalsplat}. First, the dental arch on the occlusal plane follows stable shape families—square, ovoid, and tapered—so spline/template curves can serve as global initializers or population priors, with recent AI systems automating arch-shape classification~\cite{tamayo2024dentalarch}. Second, craniofacial structures exhibit approximate bilateral symmetry; combining a mid-sagittal reference plane (MSP) with the arch midline constrains pose and lateral deviation, and recent studies review MSP construction and reliability for asymmetry analysis~\cite{ajmera2024establishment}. Third, functional cusps and occlusal contact regions provide robust cues to estimate the occlusal plane~\cite{du2025feasibility}, unifying upper/lower jaws and improving reproducibility; deep-learning pipelines on CBCT now automate landmark/plane extraction and validate measurement fidelity. Beyond these, tooth count and ordering (UR1–UR7/LR1–LR7), crown–gingiva boundary continuity, and adjacency topology form semantic–topological priors that support segmentation, instance labeling, alignment, and completion; statistical shape models (SSMs) ~\cite{kim2023developing}of whole dentitions or mandibles operationalize such priors as a “shape space”. Recent related work leverages these priors in three directions: (i) boundary-preserving 3D dental segmentation, which maintains crown–gingiva junction geometry via mesh-aware sampling and multi-view cues~\cite{xi20253d}; (ii) task-driven applications such as forensic 3D tooth identification and robust segmentation/labeling on partial-arch intraoral scans under limited coverage~\cite{mouncif20253d}; and (iii) 3D cephalometric evaluation and normative datasets~\cite{kim2025three}, which provide reusable population standards to anchor geometric alignment and quantitative assessment.

\subsection{Automated Analysis of Lateral Cephalometric Radiography}
In lateral cephalograms, the core computer-vision task is automated anatomical landmark detection and quantitative analysis. Recent systematic reviews and meta-analyses report that deep learning achieves accuracy comparable to that of expert tracing on 2D cephalograms, supporting standardized pipelines for routine measurements~\cite{serafin2023accuracy}. Methodologically, approaches span two-stage regressors and multi-scale candidate refinement to end-to-end systems, with stable errors reported on public datasets and growing evidence from prospective or clinical evaluations~\cite{wang2018automatic}. Beyond landmarking, cross-imaging 2D alignment and superimposition is increasingly integrated into workflows—for example, automatic registration between a lateral cephalogram and a facial profile photograph to jointly assess soft- and hard-tissue relations—using hierarchical contour matching or learned fusion with validated error analyses~\cite{wang2018novel}. Complementary tasks such as structure segmentation and target detection (e.g., mandibular boundary, lesions, or third-molar status) provide additional channels for automated measurement and quality control, with recent surveys highlighting robustness under low contrast and metal artifacts~\cite{khattak2025deep}. In parallel, benchmarks and protocols are being updated: new large-scale cephalometric datasets and overviews place stronger emphasis on annotation consistency, metric definitions, and stratified error reporting, improving comparability and external validity~\cite{khalid2025benchmark}. Finally, to reduce radiation exposure, emerging studies explore landmark inference without X-rays, estimating standard cephalometric points from profile photographs alone, opening avenues for multimodal fusion and streamlined care pathways~\cite{takahashi2023cephalometric}.

\begin{figure*}[!h]
    \centering
    \includegraphics[width=1\textwidth]{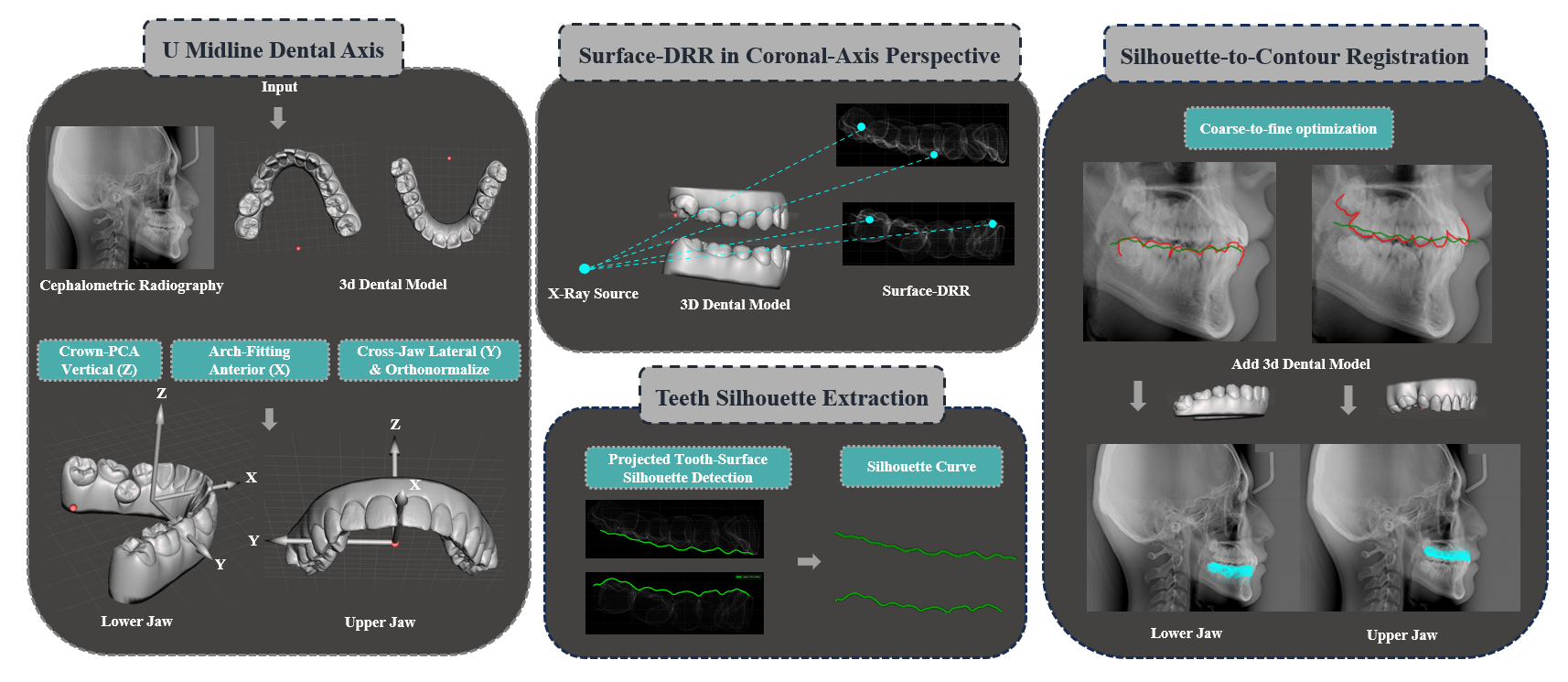}
    \vspace{-2em}
    \caption{\textbf{Overall of the proposed framework.} The system first constructs a unified U-Midline Dental Axis (UMDA) to establish a reproducible anatomical coordinate frame shared by both arches. Using this reference, intraoral scan (IOS) meshes are projected into radiograph-like views through a coronal-axis surface-based DRR with Gaussian splatting, preserving clinical magnification and geometric distortion. The resulting silhouette is then aligned to expert-annotated cephalometric contours via a 2D similarity transform optimized with a symmetric bidirectional Chamfer distance under a hierarchical coarse-to-fine schedule. This pipeline yields stable initialization, radiographically consistent projections, and accurate silhouette-to-contour registration.}
    \label{fig:pipeline}
\end{figure*}

\section{Methodology}

\subsection{U-Midline Dental Axis Construction}

We establish a unified, reproducible, and anatomically consistent 3D reference frame $\{X,Y,Z\}$ for the upper and lower dental meshes. 
The frame encodes explicit anatomical priors:
\begin{itemize}
    \item \textbf{Vertical axis ($Z$)} aligns with the dominant normal of the occlusal/crown region;
    \item \textbf{Anterior axis ($X$)} follows the U-shaped dental arch within the occlusal plane toward the incisors;
    \item \textbf{Lateral axis ($Y$)} is shared across jaws to avoid subtle yaw discrepancies from independent estimation.
\end{itemize}
This UMDA provides a stable global reference for cross-jaw alignment and statistics, a reproducible orthogonal basis for 2D projections, and a common coordinate system for quantifying occlusion, arch expansion, and inclination. Embedding the axes into each IOS model also facilitates visual inspection and batch quality control.

Let the mesh vertices be $\{v_i\}$ with normals $\{n_i\}$ and centroid 
\begin{equation}
c = \frac{1}{N} \sum_i v_i, \quad 
\text{where } \mathrm{unit}(\cdot) \text{ denotes normalization.}
\end{equation}
The UMDA is determined sequentially by estimating the vertical axis $Z$, anterior axis $X$, and shared lateral axis $Y$.

\subsubsection{Vertical Axis $Z$ via Crown Localization}
For each vertex, the angle between its surface normal and the occlusal/crown reference is computed. The top $p\%$ of vertices with highest alignment are selected as crown candidates. PCA is then applied, and the eigenvector corresponding to the smallest eigenvalue defines
\begin{equation}
Z = \mathrm{unit}\!\big(u_{\min}(C)\big),
\end{equation}
where $u_{\min}(C)$ is the eigenvector of the covariance $C$ with minimal eigenvalue. Orientation is unified by flipping $Z \!\leftarrow\! -Z$ if necessary.

\subsubsection{Anterior Axis $X$ and Initial Lateral Axis $Y_0$ via U-Shape Fitting}
Vertices are projected onto the plane orthogonal to $Z$; the dental arch is fitted using the baseline $b$ between left and right molars, with the incisal edge $f$ as an anterior reference. The midpoint $m$ on $b$ defines the 2D anterior direction:
\begin{equation}
x_{2D} = \mathrm{unit}\!\left((f - m) - ((f - m)^\top b)b\right).
\end{equation}
This vector is back-projected into 3D and orthogonalized with respect to $Z$:
\begin{equation}
X_0 = \mathrm{unit}(U x_{2D}), \quad 
Y_0 = \mathrm{unit}(Z \times X_0).
\end{equation}
Finally, $X_0$ is re-normalized via $(Y_0 \times Z)$ to ensure orthogonality and stability.

\subsubsection{Shared Lateral Axis $Y$ via Cross-Jaw Consistency}
To suppress minor asymmetries, the lateral directions from both jaws are averaged and normalized:
\begin{equation}
Y_{\mathrm{sh}} = \mathrm{unit}\!\big(Y^{(\mathrm{up})}_0 + Y^{(\mathrm{lo})}_0 \big).
\end{equation}
With fixed $Z$, the orthogonal basis is completed while maintaining $X$ consistency:
\begin{equation}
X = \mathrm{unit}(Y_{\mathrm{sh}} \times Z), \quad Y = Y_{\mathrm{sh}}.
\end{equation}
The final UMDA frame is $R = [X,Y,Z]$ with origin $o = c$.

The UMDA induces three clinically interpretable anatomical planes: the X–Y plane (occlusal plane), which is orthogonal to the Z-axis and serves as a stable basis for 2D projections; the X–Z plane (U-midline plane), which encodes midline symmetry to assess anterior–posterior and vertical deviations; and the Y–Z plane (coronal plane), which captures left–right symmetry and is useful for evaluating lateral tilt and arch inclination.

\subsection{Surface-DRR in Coronal-Axis Perspective}

After establishing the UMDA coordinate system, we further simulate the true radiographic geometry to generate lateral X-ray–like projections consistent with clinical cephalometric imaging. 
Traditional DRR are typically based on volumetric data such as CT or CBCT, where the X-ray intensity at each detector pixel $(u,v)$ is computed by integrating the attenuation coefficient $\mu(\mathbf{x})$ along the corresponding ray path $\mathcal{R}(u,v)$ according to the Beer–Lambert law:
\begin{equation}
I(u,v) = I_0 \exp\!\Big(-\!\!\int_{\mathcal{R}(u,v)} \!\mu(\mathbf{x})\,\mathrm{d}l \Big).
\end{equation}
However, intraoral scan (IOS) models represent only the tooth surfaces as triangular meshes without internal density information. 
Since our registration objective focuses on the external \emph{silhouette} rather than internal attenuation, we adopt a \textbf{surface-based DRR} formulation. 
The mesh is treated as a hollow shell of uniform density $\rho_0$, and the projected contribution of each surface element is accumulated on the detector plane to form an equivalent thickness map $L(u,v)$:
\begin{equation}
L(u,v) = \sum_{i=1}^{N} \rho_0\, a_i\, K_\sigma(u - y'_i,\, v - z'_i),
\end{equation}
where $a_i$ is the area weight of the $i$-th surface element and $K_\sigma$ is a kernel centered at $(y'_i, z'_i)$ (typically Gaussian). 
The final intensity is then derived by substituting $L(u,v)$ into the Beer–Lambert model:
\begin{equation}
I(u,v) = I_0 \exp(-\mu_0\, L(u,v)),
\end{equation}
where $\mu_0$ is a uniform effective attenuation coefficient. 
This surface-based DRR preserves the physical interpretability of radiographic formation while avoiding fictitious volumetric assumptions, emphasizing geometrically accurate outer contours.

In terms of imaging geometry, we employ a \textbf{coronal-axis perspective model} consistent with clinical lateral cephalograms. 
In the UMDA coordinate frame, $X$ denotes the left–right direction, $Y$ denotes anterior–posterior, and $Z$ denotes superior–inferior. 
The X-ray source is placed at $S = (x_s, 0, 0)$ along the $+X$ direction, and the detector plane is defined as $X = X_{\mathrm{DET}}$, such that rays propagate along the coronal axis through the dentition model. 
For a vertex $V = (x, y, z)$, the perspective projection onto the detector plane follows:
\begin{equation}
P(t) = S + t(V - S), \qquad 
t = \frac{X_{\mathrm{DET}} - x_s}{x - x_s},
\end{equation}
and the corresponding detector coordinates are
\begin{equation}
(y', z') = (t\,y,\; t\,z).
\end{equation}
This configuration explicitly models the depth-dependent magnification and geometric distortion determined by the source–object–detector (SOD/SID) geometry, yielding a radiographic appearance and spatial proportion consistent with real CR.

\paragraph{Gaussian splatting approximation.}
Because the intraoral mesh is a sparse surface point set, direct projection produces discontinuous and aliased silhouettes. 
We therefore employ a \textbf{Gaussian splatting} strategy to approximate a continuous thickness distribution:
\begin{equation}
L(u,v) = \sum_{i=1}^{N} w_i
\exp\!\Big(-\frac{(u - y'_i)^2 + (v - z'_i)^2}{2\sigma_i^2}\Big),
\end{equation}
where $w_i = \rho_0 a_i$ is the effective weight and $\sigma_i$ controls spatial smoothness. 
When used within the exponential attenuation model $I(u,v) = I_0 \exp(-\mu_0 L(u,v))$, this formulation produces radiograph-like intensity with soft gradients and realistic contours. 
Alternatively, when only geometric alignment is required, $L(u,v)$ can be directly used as a silhouette map without radiometric scaling.

\subsection{Silhouette-to-Contour Registration}

After obtaining the lateral X-ray projection of the dental IOS model meshes, we align the \emph{projected IOS model silhouette} (outer boundary of the 3D projection) to the \emph{annotated CR contour} (polyline on the radiograph) to establish geometric correspondence between the synthesized projection and the clinical CR. We formulate this as a 2D similarity transform that maps the projected IOS model silhouette/contour $\mathcal{C}_{\text{IOS model}}$ onto the CR contour $\mathcal{C}_{\text{CR}}$. For a projected vertex $\mathbf{p} = [y,z]^\top$, the transform is
\begin{equation}
\mathbf{p}' = s \, R(\theta) 
\begin{bmatrix}
y \\ -z
\end{bmatrix}
+ 
\begin{bmatrix}
t_x \\ t_y
\end{bmatrix},
\end{equation}
where $s$ is the isotropic scale, $(t_x,t_y)$ are in-plane translations, $\theta$ is the in-plane rotation, and the sign inversion on $z$ follows the radiographic convention that image $y$ increases downward.

To quantify alignment, we use the symmetric Chamfer distance between the two sets:
\begin{equation}
\begin{aligned}
\mathcal{L}_{\text{Chamfer}} 
= & \; \frac{1}{|\mathcal{C}_{\text{IOS model}}|} 
\sum_{\mathbf{p} \in \mathcal{C}_{\text{IOS model}}} 
\min_{\mathbf{q} \in \mathcal{C}_{\text{CR}}} 
\|\mathbf{p}-\mathbf{q}\|_2 \\
& + \frac{1}{|\mathcal{C}_{\text{CR}}|} 
\sum_{\mathbf{q} \in \mathcal{C}_{\text{CR}}} 
\min_{\mathbf{p} \in \mathcal{C}_{\text{IOS model}}} 
\|\mathbf{q}-\mathbf{p}\|_2 .
\end{aligned}
\end{equation}

For robustness and accuracy, we adopt a hierarchical coarse-to-fine optimization. We first initialize the similarity parameters by matching bounding boxes and centroids of $\mathcal{C}_{\text{IOS model}}$ and $\mathcal{C}_{\text{CR}}$ to obtain a scale-consistent, well-centered starting point. A \emph{coarse} stage then explores a broad parameter range to reject spurious hypotheses and accommodate inter-case variability (including jaw differences). Next, a \emph{fine} stage shrinks the bounds around the current estimate to sharpen alignment to the anatomical outline. Finally, a \emph{super-fine} stage performs sub-pixel refinement within the local basin to further reduce residuals. This multi-stage procedure balances global stability with local precision, yielding reliable, high-accuracy \emph{silhouette-to-contour} registration across cases.

\section{Experiments}
\subsection{Dataset and Annotation Policy}

We collected 34 cases; each case includes upper and lower 3D dental models and the corresponding CR. All contours and landmarks were independently annotated and reconciled by two experienced orthodontists. Tooth codes follow standard orthodontic notation: \textbf{UR = Upper Right}, \textbf{LR = Lower Right}; digits 1--7 denote central incisor, lateral incisor, canine, first premolar, second premolar, first molar, and second molar, respectively. To avoid target leakage, the training signal comes from contours only, whereas quantitative evaluation uses landmarks only.

\subsubsection{Contours for Training}
\label{subsec:train_contours}

\begin{figure}[h]
    \centering
    \includegraphics[width=0.8\linewidth]{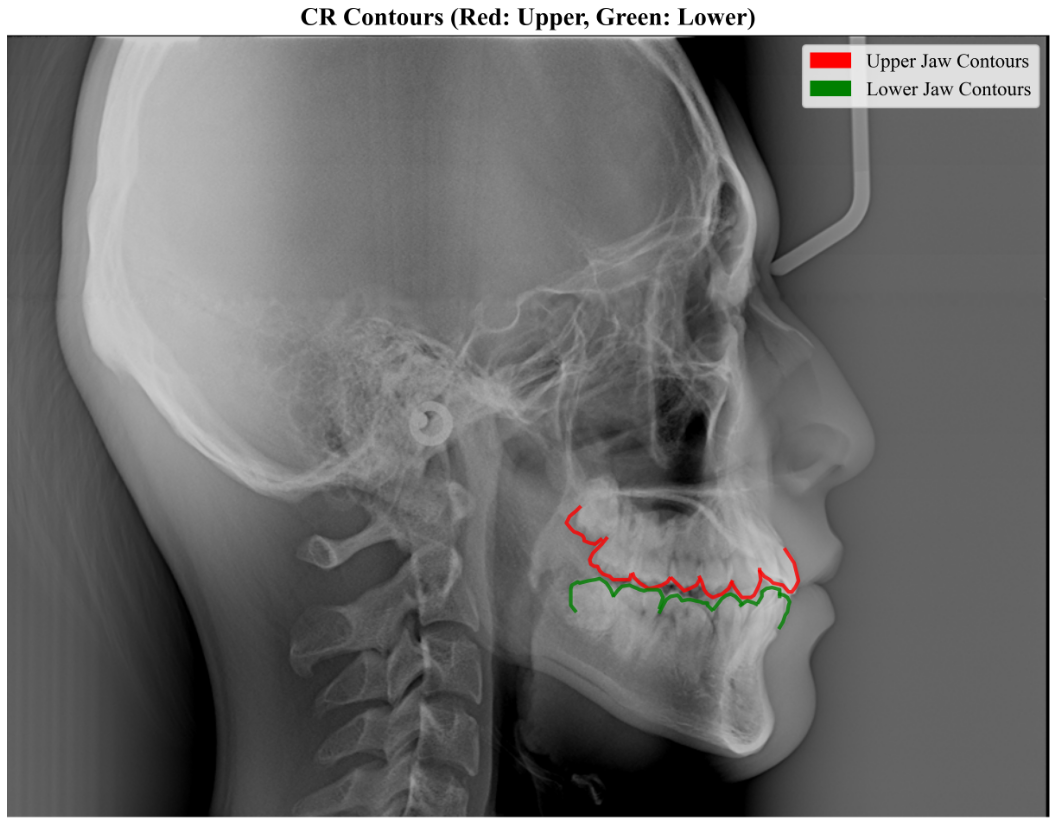} 
    \caption{Contours on the cephalometric radiograph. Upper arch shown in red and lower arch in green; each arch is formed by concatenating tooth-level labial/buccal polylines in anatomical order.}
    \label{fig:cr_contours}
\end{figure}

On the CR, the visible labial/buccal boundary of each tooth is traced as segmented polylines and concatenated, in anatomical order, into a continuous arch per jaw. The resulting upper and lower arches—illustrated in Fig.~\ref{fig:cr_contours}—are used to construct the distance field and to drive contour-to-contour alignment. Table~\ref{tab:tooth_code_map} lists the tooth-code mapping used during annotation.

\begin{table}[t]
\centering
\small
\caption{Tooth code mapping used for contour annotation (UR/LR combined).}
\label{tab:tooth_code_map}
\begin{tabularx}{\columnwidth}{@{}lX lX@{}}
\toprule
\multicolumn{2}{c}{\textbf{Upper Right (UR)}} & \multicolumn{2}{c}{\textbf{Lower Right (LR)}} \\
\cmidrule(r){1-2}\cmidrule(l){3-4}
\textbf{Code} & \textbf{Tooth (EN)} & \textbf{Code} & \textbf{Tooth (EN)} \\
\midrule
UR1 & Central incisor   & LR1 & Central incisor \\
UR2 & Lateral incisor   & LR2 & Lateral incisor \\
UR3 & Canine            & LR3 & Canine \\
UR4 & First premolar    & LR4 & First premolar \\
UR5 & Second premolar   & LR5 & Second premolar \\
UR6 & First molar       & LR6 & First molar \\
UR7 & Second molar      & LR7 & Second molar \\
\bottomrule
\end{tabularx}
\end{table}

\subsubsection{Landmarks for Evaluation}
\label{subsec:eval_landmarks}

\begin{figure}[h]
    \centering
    \includegraphics[width=1\linewidth]{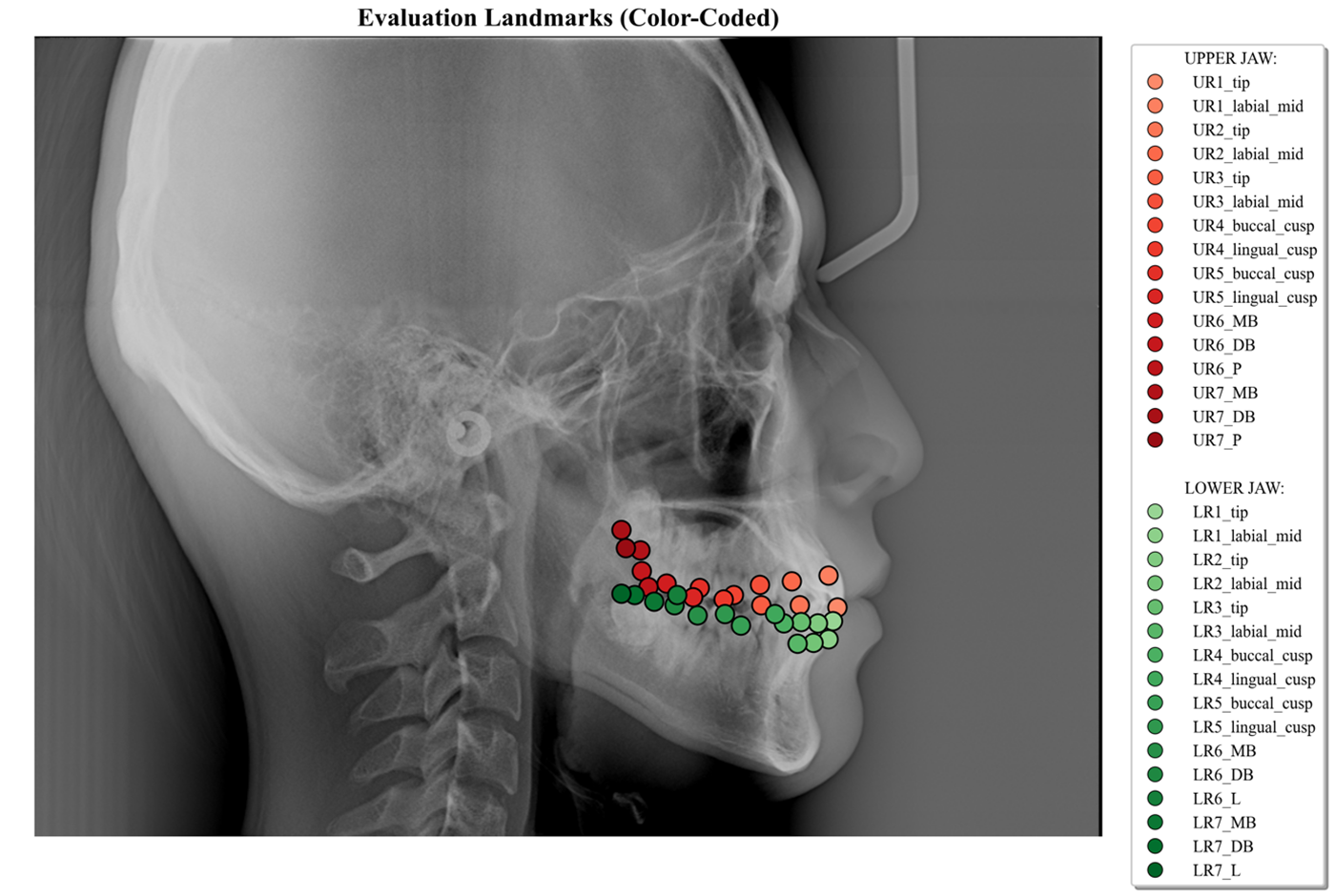}
    \vspace{0.01cm}
    \caption{Two-dimensional view displaying the corresponding anatomical landmarks on a lateral cephalometric radiograph. Landmark definitions and their corresponding codes are detailed in Table~\ref{tab:landmark_code_map}.}
    \label{fig:landmark2d}
\end{figure}

\begin{figure}[h]
    \centering
    \includegraphics[width=1\linewidth]{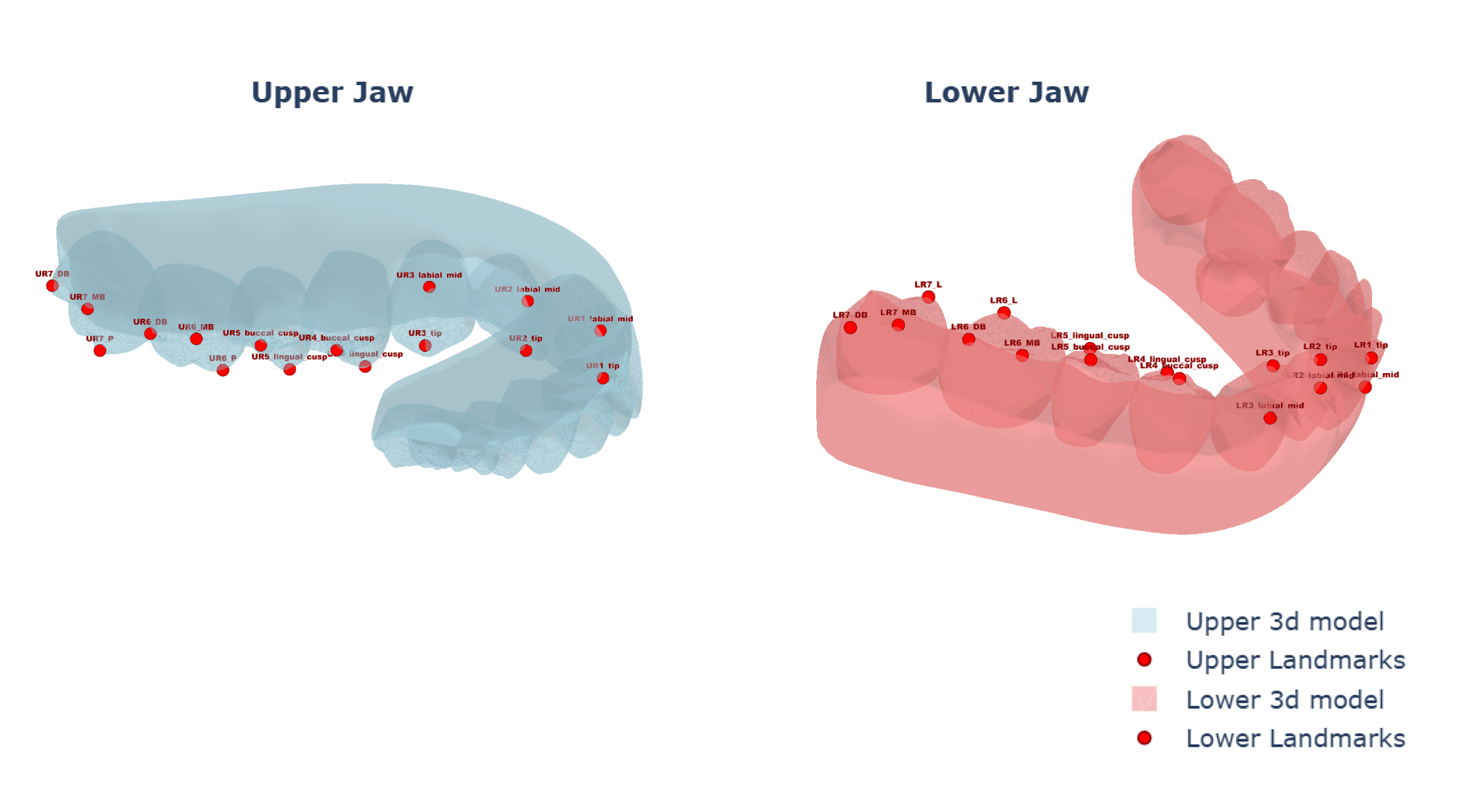}
    \vspace{0.01cm}
    \caption{Three-dimensional visualization of the upper jaw (UR, blue) and lower jaw (LR, red) dental models with overlaid anatomical landmarks (red points). The anatomical significance of the corresponding landmarks is detailed in Table~\ref{tab:landmark_code_map}.}
    \label{fig:landmark3d}
\end{figure}

Figures~\ref{fig:landmark2d} and Figures~\ref{fig:landmark3d} illustrate the evaluation landmarks in 3D and 2D, respectively. In 3D, the upper (UR, blue) and lower (LR, red) IOS model surfaces are overlaid with landmark points spanning anterior and posterior segments. In 2D, the same set is placed on the lateral CR. Landmarks are selected to be clinically meaningful, visually distinctive, and consistently visible across IOS model and CR. The corresponding codes and anatomical definitions are summarized in Table~\ref{tab:landmark_code_map}.

\emph{Selection rationale:} to cover both jaws and both anterior/posterior segments with points that are clinically meaningful, visually distinctive, and cross-modality visible. Incisal edges provide midline and vertical references; canine cusps act as robust occlusal anchors; first-molar buccal cusps serve as standard occlusal benchmarks; when visible, second-molar mesiobuccal cusps extend posterior coverage. All points must be clearly identifiable on both IOS model and CR. Table~\ref{tab:landmark_code_map} summarizes the UR/LR pairs and their anatomical definitions.

\begin{table}[t]
\centering
\small
\caption{Evaluation landmark mapping (UR/LR combined).}
\label{tab:landmark_code_map}
\begin{tabular}{@{}llp{4cm}@{}}
\toprule
\textbf{Code (UR)} & \textbf{Code (LR)} & \textbf{Anatomical definition} \\
\midrule
\addlinespace[4pt]
UR1\_tip  & LR1\_tip  & Central incisor incisal edge \\
UR2\_tip  & LR2\_tip  & Lateral incisor incisal edge \\
UR3\_cusp & LR3\_cusp & Canine cusp tip \\
UR6\_MB   & LR6\_MB   & 1st molar mesiobuccal cusp \\
UR6\_DB   & LR6\_DB   & 1st molar distobuccal cusp \\
UR7\_MB   & LR7\_MB   & 2nd molar mesiobuccal cusp  \\
\bottomrule
\end{tabular}
\end{table}

\begin{figure*}[t]
    \centering
    \includegraphics[width=\linewidth]{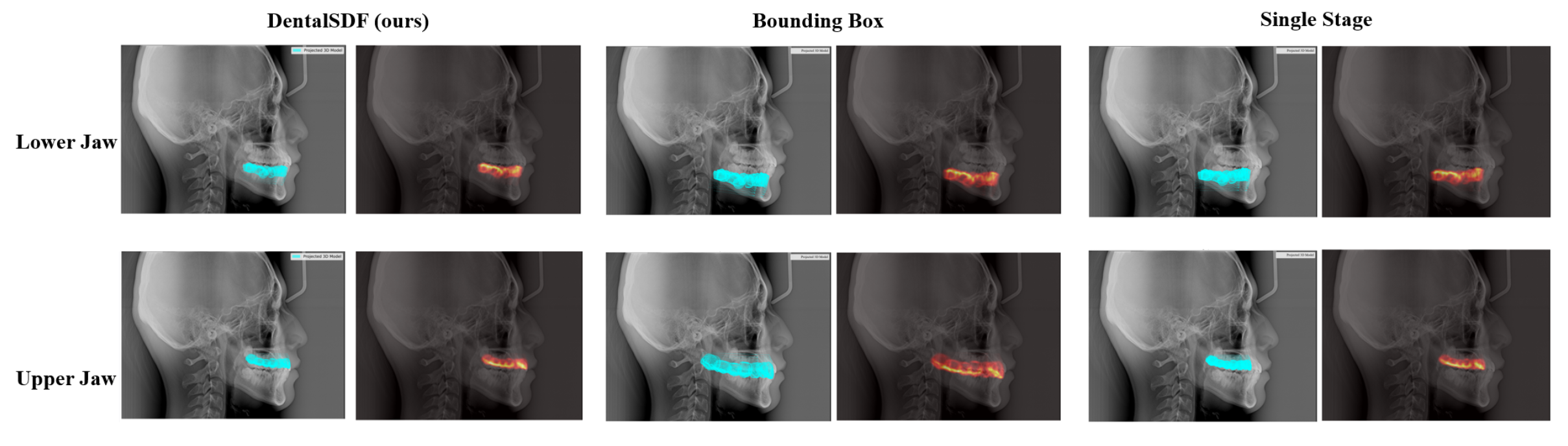}
    \vspace{0.01cm}
    \caption{Qualitative comparison on representative cases. Columns: DentalSCR (ours), Bounding Box, Single-Stage. DentalSCR produces tighter IOS model$\rightarrow$CR overlays with fewer gaps, especially at incisors/canines and posterior cusps.}
    \label{fig:baseline_vis}
\end{figure*}

\subsubsection{Per-case Data Size and Registration Parameters}
For completeness, the data dimensions used in the silhouette-guided optimization are reported. The CR image size was $703 \times 938$ pixels. The number of contour points extracted was 79 for the upper jaw and 75 for the lower jaw on the CR, which were matched against 441 and 439 points from the IOS model contours for the upper and lower jaw, respectively.

\subsection{Evaluation metrics}

Let $\mathcal{C}_{\text{3D}}=\{\,\mathbf{p}_i\in\mathbb{R}^2\,\}_{i=1}^{N}$ be the projected \textit{IOS model} silhouette points and
$\mathcal{C}_{\text{CR}}=\{\,\mathbf{q}_j\in\mathbb{R}^2\,\}_{j=1}^{M}$ the annotated CR contour points.
Distances are Euclidean in pixels and lower is better. For a point set $\mathcal{A}$ and $\mathcal{B}$, define the point-to-set distance
\begin{equation}
d(\mathbf{x},\mathcal{B}) \;=\; \min_{\mathbf{y}\in\mathcal{B}} \|\mathbf{x}-\mathbf{y}\|_2 .
\end{equation}

\subsubsection{One-sided mean distances}
We report the two directed means to diagnose asymmetric coverage:
\begin{equation}
\begin{aligned}
\overline{D}\big(\mathcal{C}_{\text{3D}}\!\to\!\mathcal{C}_{\text{CR}}\big)
&= \frac{1}{N}\sum_{i=1}^{N} d\!\left(\mathbf{p}_i,\mathcal{C}_{\text{CR}}\right),\\
\overline{D}\big(\mathcal{C}_{\text{CR}}\!\to\!\mathcal{C}_{\text{3D}}\big)
&= \frac{1}{M}\sum_{j=1}^{M} d\!\left(\mathbf{q}_j,\mathcal{C}_{\text{3D}}\right).
\end{aligned}
\end{equation}

\subsubsection{Chamfer distance (bi-directional mean)}
We use the symmetric (averaged) Chamfer distance:
\begin{equation}
\mathrm{CD}\!\left(\mathcal{C}_{\text{3D}},\mathcal{C}_{\text{CR}}\right)
= \frac{1}{2}\Big[
\overline{D}\big(\mathcal{C}_{\text{3D}}\!\to\!\mathcal{C}_{\text{CR}}\big)
+
\overline{D}\big(\mathcal{C}_{\text{CR}}\!\to\!\mathcal{C}_{\text{3D}}\big)
\Big].
\end{equation}
(Equivalently, some works sum the two directed means; we report their average.)

\subsubsection{Hausdorff distance}
The directed Hausdorff distance is
\begin{equation}
h\big(\mathcal{A}\!\to\!\mathcal{B}\big)
= \max_{\mathbf{x}\in\mathcal{A}} d(\mathbf{x},\mathcal{B}),
\end{equation}
and the symmetric Hausdorff distance is
\begin{equation}
\mathrm{H}\!\left(\mathcal{C}_{\text{3D}},\mathcal{C}_{\text{CR}}\right)
= \max\!\Big\{
h\big(\mathcal{C}_{\text{3D}}\!\to\!\mathcal{C}_{\text{CR}}\big),\;
h\big(\mathcal{C}_{\text{CR}}\!\to\!\mathcal{C}_{\text{3D}}\big)
\Big\}.
\end{equation}

\section{Quantitative and qualitative results}

We compare three settings under the unified UMDA frame using landmarks as the primary metric: Bounding Box (initial alignment only), Single-Stage (80 iterations), and DentalSCR with a three-stage schedule. As summarized in Table~\ref{tab:baseline_landmark_wide}, DentalSCR consistently lowers landmark errors across jaws. On the upper jaw, the mean decreases from 43.508 (Bounding Box) and 54.567 (Single-Stage) to 17.546, with RMSE dropping to 20.339 and dispersion narrowing (std 10.288). On the lower jaw, the mean reduces from 44.887/49.625 to 40.617 with RMSE 42.853. Stratifying by tooth type (Inc.\&Can.: Incisors and Canines; PreM: Premolars) shows that posterior structures benefit the most on the upper jaw (molars mean from 70.539 to 14.513), while the lower molars also improve to 24.652; by contrast, the lower anterior segment remains relatively harder (incisors \& canines mean 49.644), reflecting thinner silhouettes and stronger curvature on lateral projections.

\begin{table}[t]
\centering
\small
\setlength{\tabcolsep}{3.5pt} 
\caption{Baseline comparison with \emph{landmark-based} evaluation (pixels). Values are rounded to three decimals.}
\label{tab:baseline_landmark_wide}
\begin{tabular}{@{}lcccccc@{}}
\toprule
\multicolumn{7}{c}{\textbf{Upper Jaw}} \\
\midrule
Method & Mean & RMSE & Std & Inc.\&Can. & PreM & Molars \\
\midrule
Bounding Box & 43.508 & 50.734 & 26.094 & 18.405 & 40.618 & 70.539 \\
Single-Stage & 54.567 & 58.666 & 21.546 & 35.163 & 51.224 & 76.199 \\
DentalSCR & 17.546 & 20.339 & 10.288 & 23.472 & 13.205 & 14.513 \\
\midrule
\multicolumn{7}{c}{\textbf{Lower Jaw}} \\
\midrule
Method & Mean & RMSE & Std & Inc.\&Can. & PreM & Molars \\
\midrule
Bounding Box & 44.887 & 46.859 & 13.451 & 29.939 & 54.100 & 53.693 \\
Single-Stage & 49.625 & 50.662 & 10.199 & 38.577 & 58.672 & 54.640 \\
DentalSCR & 40.617 & 42.853 & 13.660 & 49.644 & 51.025 & 24.652 \\
\bottomrule
\end{tabular}%
\end{table}

To contextualize the final fit quality at the curve level, Table~\ref{tab:baseline_contour_single_vertical} reports contour-distance metrics for DentalSCR only. The bi-directional Chamfer is 8.397 for the upper jaw and 4.055 for the lower jaw, indicating tight two-way coverage of the CR silhouettes. The Hausdorff distances (45.714 upper; 22.045 lower) suggest residual worst-case discrepancies primarily in thin or highly curved regions. The single-sided means (6.082 IOS model$\rightarrow$CR vs.\ 10.713 CR$\rightarrow$IOS model on the upper jaw) further indicate mild under-coverage of certain annotated edges by the projected IOS model, which aligns with the larger upper-jaw Hausdorff. Qualitatively, visual inspection (not shown) confirms that Single-Stage aligns the bulk but leaves small gaps around incisal edges and canine tips, whereas the three-stage DentalSCR reduces these residuals and yields cleaner overlap at posterior cusps.

\begin{table}[t]
\centering
\small
\caption{Contour-distance metrics for the final DentalSCR alignment (pixels). Values are rounded to three decimals.}
\label{tab:baseline_contour_single_vertical}
\begin{tabular}{@{}lcc@{}}
\toprule
\textbf{Metric} & \textbf{Upper Jaw} & \textbf{Lower Jaw} \\
\midrule
Chamfer distance (bi-directional mean) & 8.397 & 4.055 \\
Hausdorff distance                      & 45.714 & 22.045 \\
IOS $\rightarrow$ CR mean distance      & 6.082 & 3.585 \\
CR $\rightarrow$ IOS mean distance      & 10.713 & 4.526 \\
\bottomrule
\end{tabular}
\end{table}

\subsection{Ablation studies}

We evaluate each component's contribution via landmark-based evaluation on both jaws (Table~\ref{tab:ablation_landmark_single}). Removing the UMDA coordinate system causes the most significant degradation, with mean errors reaching 49.472 (upper) and 80.423 (lower), and substantially increased dispersion on the lower jaw (std 37.125), highlighting the necessity of a stable cross-case, cross-jaw frame. Eliminating the coarse-to-fine schedule also reduces accuracy (upper 54.567; lower 49.625), indicating that single-stage optimization tends to stall with residual misalignment even when variability appears moderate. Replacing symmetric Chamfer distance with a one-sided variant raises the means to 42.089 (upper) and 48.871 (lower), demonstrating that balanced bidirectional coverage is essential to prevent systematic silhouette under- or over-coverage. Among all variants, DentalSCR achieves the lowest means (17.546 upper; 40.617 lower) and RMSEs (20.339; 42.853), confirming that integrating UMDA, symmetric Chamfer, and the three-stage schedule is crucial for consistent landmark alignment.

\begin{table}[t]
\centering
\small
\caption{Ablation results with \emph{landmark-based} evaluation (pixels). Upper and lower jaws are shown side-by-side; values are rounded to three decimals.}
\label{tab:ablation_landmark_single}
\resizebox{\columnwidth}{!}{%
\begin{tabular}{@{}lccc|ccc@{}}
\toprule
 & \multicolumn{3}{c|}{Upper Jaw (landmarks)} & \multicolumn{3}{c}{Lower Jaw (landmarks)} \\
\cmidrule(lr){2-4}\cmidrule(lr){5-7}
Config & Mean & RMSE & Std & Mean & RMSE & Std \\
\midrule
DentalSCR (Ours)   & 17.546 & 20.339 & 10.288 & 40.617 & 42.853 & 13.660 \\
w/o UMDA           & 49.472 & 52.328 & 17.050 & 80.423 & 88.578 & 37.125 \\
w/o Bi-Chamfer     & 42.089 & 46.296 & 19.285 & 48.871 & 50.004 & 10.588 \\
w/o Coarse-to-Fine & 54.567 & 58.666 & 21.546 & 49.625 & 50.662 & 10.199 \\
\bottomrule
\end{tabular}%
}
\end{table}

\section{Conclusion and Future Work}

We presented \emph{DentalSCR}, a silhouette–distance field framework for registering 3D dental models to lateral cephalograms. It integrates a reproducible U-midline Dental Axis (UMDA), perspective-consistent rendering, and a symmetric Chamfer objective within a coarse-to-fine schedule. On expert-annotated cases, our method reduced landmark errors, improved posterior alignment, and achieved interpretable curve-level agreement. Future work will enhance robustness for thin structures with orientation/curvature cues, extend to multi-view settings for better depth constraints, and incorporate uncertainty-aware weighting for noisy contours, validated through multi-center studies.

\section{Acknowledgments}
This research is funded by the Postgraduate Research Scholarship (PGRS) at Xi’an Jiaotong-Liverpool University, contract number TW5A2312001 and FOS2104JP08.

\bibliographystyle{IEEEtran}
\bibliography{reference}

@article{maintz1998survey,
  title={A survey of medical image registration},
  author={Maintz, JB Antoine and Viergever, Max A},
  journal={Medical image analysis},
  volume={2},
  number={1},
  pages={1--36},
  year={1998},
  publisher={Elsevier}
}

@article{zitova2003image,
  title={Image registration methods: a survey},
  author={Zitova, Barbara and Flusser, Jan},
  journal={Image and vision computing},
  volume={21},
  number={11},
  pages={977--1000},
  year={2003},
  publisher={Elsevier}
}

@article{wells1996multi,
  title={Multi-modal volume registration by maximization of mutual information},
  author={Wells III, William M and Viola, Paul and Atsumi, Hideki and Nakajima, Shin and Kikinis, Ron},
  journal={Medical image analysis},
  volume={1},
  number={1},
  pages={35--51},
  year={1996},
  publisher={Elsevier}
}

@article{penney1998comparison,
  title={A comparison of similarity measures for use in 2-D-3-D medical image registration},
  author={Penney, Graeme P and Weese, J{\"u}rgen and Little, John A and Desmedt, Paul and Hill, Derek LG and others},
  journal={IEEE transactions on medical imaging},
  volume={17},
  number={4},
  pages={586--595},
  year={1998},
  publisher={IEEE}
}

@article{maes2002multimodality,
  title={Multimodality image registration by maximization of mutual information},
  author={Maes, Frederik and Collignon, Andre and Vandermeulen, Dirk and Marchal, Guy and Suetens, Paul},
  journal={IEEE transactions on Medical Imaging},
  volume={16},
  number={2},
  pages={187--198},
  year={2002},
  publisher={IEEE}
}

@article{viola1997alignment,
  title={Alignment by maximization of mutual information},
  author={Viola, Paul and Wells III, William M},
  journal={International journal of computer vision},
  volume={24},
  number={2},
  pages={137--154},
  year={1997},
  publisher={Springer}
}

@inproceedings{jiang1992image,
  title={Image registration of multimodality 3D medical images by chamfer matching},
  author={Jiang, Hongjian and Holton, Kerrie S and Robb, Richard A},
  booktitle={Biomedical Image Processing and Three-Dimensional Microscopy},
  volume={1660},
  pages={356--366},
  year={1992},
  organization={SPIE}
}

@article{swinehart1962beer,
  title={The beer-lambert law},
  author={Swinehart, Donald F},
  journal={Journal of chemical education},
  volume={39},
  number={7},
  pages={333},
  year={1962},
  publisher={ACS Publications}
}

@article{rino2013evaluation,
  title={Evaluation of radiographic magnification in lateral cephalograms obtained with different X-ray devices: experimental study in human dry skull},
  author={Rino Neto, Jos{\'e} and Paiva, Jo{\~a}o Batista de and Queiroz, Gilberto Vilanova and Attizzani, Miguel Ferragut and Miasiro Junior, Hiroshi},
  journal={Dental Press Journal of Orthodontics},
  volume={18},
  pages={17e1--17e7},
  year={2013},
  publisher={SciELO Brasil}
}

@article{levine2020drrgenerator,
  title={DRRGenerator: a three-dimensional slicer extension for the rapid and easy development of digitally reconstructed radiographs},
  author={Levine, Lance and Levine, Marc},
  journal={Journal of Clinical Imaging Science},
  volume={10},
  pages={69},
  year={2020}
}

@inproceedings{unberath2018deepdrr,
  title={DeepDRR--a catalyst for machine learning in fluoroscopy-guided procedures},
  author={Unberath, Mathias and Zaech, Jan-Nico and Lee, Sing Chun and Bier, Bastian and Fotouhi, Javad and Armand, Mehran and Navab, Nassir},
  booktitle={International conference on medical image computing and computer-assisted intervention},
  pages={98--106},
  year={2018},
  organization={Springer}
}

@article{huttenlocher2002comparing,
  title={Comparing images using the Hausdorff distance},
  author={Huttenlocher, Daniel P and Klanderman, Gregory A. and Rucklidge, William J},
  journal={IEEE Transactions on pattern analysis and machine intelligence},
  volume={15},
  number={9},
  pages={850--863},
  year={2002},
  publisher={IEEE}
}

@inproceedings{besl1992method,
  title={Method for registration of 3-D shapes},
  author={Besl, Paul J and McKay, Neil D},
  booktitle={Sensor fusion IV: control paradigms and data structures},
  volume={1611},
  pages={586--606},
  year={1992},
  organization={Spie}
}

@article{myronenko2010point,
  title={Point set registration: Coherent point drift},
  author={Myronenko, Andriy and Song, Xubo},
  journal={IEEE transactions on pattern analysis and machine intelligence},
  volume={32},
  number={12},
  pages={2262--2275},
  year={2010},
  publisher={IEEE}
}

@article{serafin2023accuracy,
  title={Accuracy of automated 3D cephalometric landmarks by deep learning algorithms: systematic review and meta-analysis},
  author={Serafin, Marco and Baldini, Benedetta and Cabitza, Federico and Carrafiello, Gianpaolo and Baselli, Giuseppe and Del Fabbro, Massimo and Sforza, Chiarella and Caprioglio, Alberto and Tartaglia, Gianluca M},
  journal={La radiologia medica},
  volume={128},
  number={5},
  pages={544--555},
  year={2023},
  publisher={Springer}
}

@article{hendrickx2024can,
  title={Can artificial intelligence-driven cephalometric analysis replace manual tracing? A systematic review and meta-analysis},
  author={Hendrickx, Julie and Gracea, Rellyca Sola and Vanheers, Michiel and Winderickx, Nicolas and Preda, Flavia and Shujaat, Sohaib and Jacobs, Reinhilde},
  journal={European Journal of Orthodontics},
  volume={46},
  number={4},
  pages={cjae029},
  year={2024},
  publisher={Oxford University Press UK}
}

@article{lee2024comparative,
  title={A comparative study of deep learning and manual methods for identifying anatomical landmarks through cephalometry and cone-beam computed tomography: A systematic review and meta-analysis},
  author={Lee, Yoonji and Pyeon, Jeong-Hye and Han, Sung-Hoon and Kim, Na Jin and Park, Won-Jong and Park, Jun-Beom},
  journal={Applied Sciences},
  volume={14},
  number={16},
  pages={7342},
  year={2024},
  publisher={MDPI}
}

@article{wang2018automatic,
  title={Automatic analysis of lateral cephalograms based on multiresolution decision tree regression voting},
  author={Wang, Shumeng and Li, Huiqi and Li, Jiazhi and Zhang, Yanjun and Zou, Bingshuang},
  journal={Journal of healthcare engineering},
  volume={2018},
  number={1},
  pages={1797502},
  year={2018},
  publisher={Wiley Online Library}
}

@article{wang2018novel,
  title={A novel contour-based registration of lateral cephalogram and profile photograph},
  author={Wang, Shumeng and Li, Huiqi and Zou, Bingshuang and Zhang, Wanjun},
  journal={Computerized Medical Imaging and Graphics},
  volume={63},
  pages={9--23},
  year={2018},
  publisher={Elsevier}
}

@inproceedings{khattak2025deep,
  title={Deep Learning Applications in Dental Image-Based Diagnostics: A Systematic Review},
  author={Khattak, Osama and Hashem, Ahmed Shawkat and Alqarni, Mohammed Saad and Almufarrij, Raha Ahmed Shamikh and Siddiqui, Amna Yusuf and Anis, Rabia and Ahmad, Shahzad and Fareed, Muhammad Amber and Alothmani, Osama Shujaa and Alkhershawy, Lama Habis Samah and others},
  booktitle={Healthcare},
  volume={13},
  number={12},
  pages={1466},
  year={2025},
  organization={MDPI}
}

@article{khalid2025benchmark,
  title={A Benchmark Dataset for Automatic Cephalometric Landmark Detection and CVM Stage Classification},
  author={Khalid, Muhammad Anwaar and Zulfiqar, Kanwal and Bashir, Ulfat and Shaheen, Areeba and Iqbal, Rida and Rizwan, Zarnab and Rizwan, Ghina and Fraz, Muhammad Moazam},
  journal={Scientific Data},
  volume={12},
  number={1},
  pages={1336},
  year={2025},
  publisher={Nature Publishing Group UK London}
}

@article{takahashi2023cephalometric,
  title={Cephalometric landmark detection without X-rays combining coordinate regression and heatmap regression},
  author={Takahashi, Kaisei and Shimamura, Yui and Tachiki, Chie and Nishii, Yasushi and Hagiwara, Masafumi},
  journal={Scientific reports},
  volume={13},
  number={1},
  pages={20011},
  year={2023},
  publisher={Nature Publishing Group UK London}
}

@article{sun2024medical,
  title={Medical image registration via neural fields},
  author={Sun, Shanlin and Han, Kun and You, Chenyu and Tang, Hao and Kong, Deying and Naushad, Junayed and Yan, Xiangyi and Ma, Haoyu and Khosravi, Pooya and Duncan, James S and others},
  journal={Medical Image Analysis},
  volume={97},
  pages={103249},
  year={2024},
  publisher={Elsevier}
}

@article{yang2024accurate,
  title={Accurate and robust registration of low overlapping point clouds},
  author={Yang, Jieyin and Zhao, Mingyang and Wu, Yingrui and Jia, Xiaohong},
  journal={Computers \& Graphics},
  volume={118},
  pages={146--160},
  year={2024},
  publisher={Elsevier}
}

@inproceedings{gopalakrishnan2024intraoperative,
  title={Intraoperative 2D/3D image registration via differentiable X-ray rendering},
  author={Gopalakrishnan, Vivek and Dey, Neel and Golland, Polina},
  booktitle={Proceedings of the IEEE/CVF Conference on Computer Vision and Pattern Recognition},
  pages={11662--11672},
  year={2024}
}

@article{leskovar2025comparison,
  title={Comparison of global and local optimization methods for intensity-based 2D--3D registration},
  author={Leskovar, Marko and Heyland, Mark and Trepczynski, Adam and Zachow, Stefan},
  journal={Computers in Biology and Medicine},
  volume={186},
  pages={109574},
  year={2025},
  publisher={Elsevier}
}

@article{tamayo2024dentalarch,
  title={Dentalarch: Ai-based arch shape detection in orthodontics},
  author={Tamayo-Quintero, JD and G{\'o}mez-Mendoza, JB and Guevara-P{\'e}rez, SV},
  journal={Applied Sciences},
  volume={14},
  number={6},
  pages={2567},
  year={2024},
  publisher={MDPI}
}

@article{ajmera2024establishment,
  title={Establishment of the mid-sagittal reference plane for three-dimensional assessment of facial asymmetry: a systematic review: establishment of the mid-sagittal reference plane: a systematic review},
  author={Ajmera, Deepal Haresh and Singh, Pradeep and Leung, Yiu Yan and Khambay, Balvinder S and Gu, Min},
  journal={Clinical Oral Investigations},
  volume={28},
  number={4},
  pages={242},
  year={2024},
  publisher={Springer}
}

@article{du2025feasibility,
  title={Feasibility of occlusal plane in predicting the changes in anteroposterior mandibular position: a comprehensive analysis using deep learning-based three-dimensional models},
  author={Du, Bingran and Li, Kaichen and Shen, Zhiling and Cheng, Yihang and Yu, Jiayan and Pan, Yaopeng and Huang, Ziyan and Hu, Fei and Rausch-Fan, Xiaohui and Zhu, Yuanpeng and others},
  journal={BMC Oral Health},
  volume={25},
  number={1},
  pages={42},
  year={2025},
  publisher={Springer}
}

@article{kim2023developing,
  title={Developing a three-dimensional statistical shape model of normal dentition using an automated algorithm and normal samples},
  author={Kim, Hwee-Ho and Choi, Sieun and Chang, Young-Il and Yi, Won-Jin and Ahn, Sug-Joon},
  journal={Clinical Oral Investigations},
  volume={27},
  number={2},
  pages={759--772},
  year={2023},
  publisher={Springer}

}

@article{mouncif20253d,
  title={3D tooth identification for forensic dentistry using deep learning},
  author={Mouncif, Hamza and Kassimi, Amine and Bertin Gardelle, Thierry and Tairi, Hamid and Riffi, Jamal},
  journal={BMC Oral Health},
  volume={25},
  number={1},
  pages={665},
  year={2025},
  publisher={Springer}
}

@article{kim2025three,
  title={Three-dimensional cephalometric evaluation of the craniofacial morphology in Korean population utilizing cone-beam computed tomography},
  author={Kim, Dong-Hyun and Li, YuWen and Lee, Kyungmin Clara},
  journal={Korean Journal of Orthodontics},
  volume={55},
  number={4},
  pages={254--265},
  year={2025},
  publisher={Korean Association of Orthodontists}
}

@article{miao2025dentalsplat,
  title={DentalSplat: Dental Occlusion Novel View Synthesis from Sparse Intra-Oral Photographs},
  author={Miao, Yiyi and Wu, Taoyu and Chen, Tong and Li, Sihao and Jiang, Ji and Yang, Youpeng and Stefanidis, Angelos and Yu, Limin and Su, Jionglong},
  journal={arXiv preprint arXiv:2511.03099},
  year={2025}
}
\end{document}